\documentclass{article}

\usepackage[numbers]{natbib}
\usepackage{multirow}
\usepackage{caption}
\usepackage{subcaption}

\usepackage[toc,page]{appendix}

\usepackage[final]{nips_2018}

\usepackage[utf8]{inputenc} 
\usepackage[T1]{fontenc}    
\usepackage{hyperref}       
\usepackage{cleveref}
\usepackage{url}            
\usepackage{booktabs}       
\usepackage{amsfonts}       
\usepackage{nicefrac}       
\usepackage{microtype}      
\usepackage{graphicx}

\title{Predicting Diabetes Disease Evolution Using Financial Records and Recurrent Neural Networks}

\author{
  Rafael T. Sousa\\
  Universidade Federal de Goiás\\
  \texttt{rafaelts777@gmail.com}\\
  \And
  Lucas A. Pereira\\
  Universidade Federal de Goiás\\
  \texttt{apereiral@outlook.com}\\
  \And
  Anderson S. Soares\\
  Universidade Federal de Goiás\\
  \texttt{engsoares@gmail.com}\\
}

\begin{document}

\maketitle

\begin{abstract}

Managing patients with chronic diseases is a major and growing healthcare challenge in several countries. A chronic condition, such as diabetes, is an illness that lasts a long time and does not go away, and often leads to the patient's health gradually getting worse. While recent works involve raw electronic health record (EHR) from hospitals, this work uses only financial records from health plan providers to predict diabetes disease evolution with a self-attentive recurrent neural network. The use of financial data is due to the possibility of being an interface to international standards, as the records standard encodes medical procedures. The main goal was to assess high risk diabetics, so we predict records related to diabetes acute complications such as amputations and debridements, revascularization and hemodialysis. Our work succeeds to anticipate complications between 60 to 240 days with an area under ROC curve ranging from 0.81 to 0.94. In this paper we describe the first half of a work-in-progress developed within a health plan provider with ROC curve ranging from 0.81 to 0.83. This assessment will give healthcare providers the chance to intervene earlier and head off hospitalizations. We are aiming to deliver personalized predictions and personalized recommendations to individual patients, with the goal of improving outcomes and reducing costs

\end{abstract}

\section{Introduction}

The last World Health Organization (WHO) diabetes report \cite{world2016global} point that the number of people with diabetes raised from 108 million in 1980 to 422 million in 2014, resulting on a global prevalence of 8.5\% among adults over 18 years of age. The estimated number of death directly caused by diabetes in 2015 was 1.6 million\cite{mathers2006projections}. According to the International Diabetes Federation, Brazil ranks fourth among countries with the highest number of diabetics, about 12.4 million in 2017 \cite{bertoldi2013epidemiology,costa2017burden,IDF2017}.


Prevention and management of diabetic population is one of the major challenges for health companies. These tasks can not only improve outcomes for patients but also substantially balance healthcare spending. As the disease progresses it can damage the heart, blood vessels, eyes, kidneys and nerves. It also increases the risk of heart diseases and strokes. The blood vessels damages combined with deterioration of the nervous system can result in foot ulcers, infection and the eventual need for limp amputation. In the same way, these damages can also make diabetes the main cause of kidney failure.

According to World Health Organization report, it is possible to prevent diabetes progress, however, efficient tools are required to identify and assess high-risk groups\cite{world2016global}. In this way, machine learning can come up with personalized predictions and recommendations significantly strengthening prevention efforts. We want to describe an approach to assess the complication risk as a diabetes progress index.

Evaluate and assess patient outcomes is a complex problem and some authors have been using machine learning in an attempt to solve it \cite{Pandey:2017:IRA:3079452.3079501,jin2018predicting,choi2016doctor,dagliati2017machine}. Their approach models the disease progression based on Eletronic Health Records (EHR) of past patients. We can highlight some remarkable works like Choi's Doctor AI \cite{choi2016doctor} which performs a multilabel prediction with Recurrent Neural Networks (RNN) applied to EHR. The results on a large real-world dataset achieved 79,58\% recall@30 and report that the modelling not only mimics the predictive power of human doctor, but also provides clinically meaningful diagnostics. Another work \cite{dagliati2017machine} presented a diabetes complication predicting technique using classic machine learning algorithms, which achieved an average 0.75 of area under ROC curve in the prediction of retinopathy, nephropathy and neuropathy.

Both works cited and others \cite{Pandey:2017:IRA:3079452.3079501,jin2018predicting} relies on a complete electronic report about patients, like demographic data, body mass index, habits and results of exams and laboratory tests. Based on the scenario of most developing countries, in several healthcare companies electronic medical records and exam results are usually either unavailable in digital format or too heterogeneous to be integrated. In this work, we propose the use of financial records as an alternative since it is, among the available data, more reliable and easy to gather. However, financial records are in a different data domain, with more sparsity and irrelevant records. To deal with it, we are also proposing a recurrent model based on recent advances on natural language process.

Inspired by the Brazilian data availability and the difference in data domain, we propose a recurrent neural network architecture to predict diabetes complications through financial records of health plan providers. We assess the risk of diabetic complications as an index to diabetes progress. In the next sections we will expound details of the data used, method and result obtained, limitations, conclusions and future works.

\section{Financial Records}

In this work we used a dataset of financial records from a Brazilian health plan provider. The last five years of data from their customers were used. There is around $7,000,000$ unique individuals with $327$ million records.

The records follow a national standard from Brazilian National Regulatory Agency for Private Health Insurance and Plans (ANS), which is called TUSS (\emph{Terminologia Unificada em Sa\'ude Suplementar} - Supplementary Healthcare Unified Terminology). The TUSS terminology was created in 2010 by Brazilian government to standardize the payment to healthcare companies, based on what services will be performed or have already been, and the exchange of information between health plan providers. The terminology have unique codes for: medical procedures, hospital and clinics rates, materials, medicines and special materials like orthoses and prostheses. Some examples are on Table 1.

\begin{table}[htb]
\centering
\begin{minipage}{.5\textwidth}
\centering
\caption{TUSS interoperability example}
\label{tuss-table1}
\begin{tabular}{c|l}
Terminology                                 &  Hemoglobin A1c \\ \hline
TUSS                                        &  40302733 \\
CPT                                         &  83037 \\
LOINC                                       &  4548-4 \\
SNOMED-CT                                   &  43396009
\end{tabular}
\end{minipage}%
\begin{minipage}{.5\textwidth}
\centering
\caption{TUSS Codes Examples}
\label{tuss-table2}
\begin{tabular}{c|l}
TUSS    & Description \\ \hline
10101012 &  Appointment \\
10101039 &  Emergency Room Service \\
40901475 &  Color Doppler of Aorta and Iliac \\
90019415 &  Dipyrone 500mg
\end{tabular}
\end{minipage}
\end{table}

TUSS encoding can be used as an interface to other procedures or medicine standards like the American Healthcare Common Procedure Coding System or the International Classification of Procedures in Medicine, providing an international interoperability.

As we are using financial records, complications were identified through related records. Some of the main diabetes complications is kidney failure and cardiovascular diseases, so we focused on three kinds of records: (i) Amputations and debridements; (ii) Revascularization and angioplasty; and (iii) Hemodialysis.

\section{Proposed Model}

The proposed model is mainly inspired by the self-attention sentence embedding proposed by \cite{DBLP:journals/corr/LinFSYXZB17}. The idea behind using self-attention is due to its capacity of select the most relevant sentences. We believe this can be useful to select relevant records and sequence of records among a long input with other possibly irrelevant records, like basic hospital materials, as needles and serum. 

The model have an embedding layer connected to a bidirectional Long Short-Term Memory (LSTM) with the self-attention mechanism followed by two fully-connected layers. 

The input embedding layer is pre-trained with Word2Vec skipgrams \cite{mikolov2013distributed} extracted from the whole dataset. As each code is unique as a word we expect to create a vector representation to deal with proprieties that the records have in common to natural language, like: synonyms, there is codes from the same drug but with different doses; antonyms, some drugs have the opposite effect; words combination, some protocols have a standard sequence of exams and drugs and it can be represented as the sum of different codes. This kind of input pre-processing was already done successfully by other works like \cite{choi2016doctor, bajor2016predicting, lipton2015learning} as a way to extract information from the whole dataset and also deal with the high number of different codes, in our case we have around $150,000$ different codes.

The main idea is to use the network to assess the likelihood that the diabetic will have a complication after a time gap.

\subsection{Experiments}

To asses prediction capacity we made tests with time gaps of 60, 120, 180 and 240 days as prediction windows. The network input is restricted to the last twelve months of records, with a maximum limit of 500 records due to the vanish gradient problem with LSTMs and the computational cost, and a minimum of 40 records to ensure enough data to make a decision and to exclude individuals with too few records. For each diabetic who had a complication we extracted one input sequence before the first record of complication, and for those who do not had complications we extracted one sequence randomly selected. 

As we do not have any diagnosis to confirm who is a diabetic or not and the company do not requires the healthcare providers to inform a diagnosis or an ICD (International Classification of Diseases) code, we define a basic filter to find the greatest possible number of diabetics. As the glycated haemoglobin is define by WHO as the standard diabetes diagnosis test \cite{who2011glycated},  all individuals who had at least two glycated haemoglobin test in less then a year were considered a possible diabetic. The filter used selected $100,000$ individuals which is coherent with the diabetes statistics in Brazil ($8.9\%$ of the country's population according to government data), since we have around two millions individuals with more than one year of data we have $5\%$ considered diabetic.

The low prevalence of complications makes the dataset imbalanced. Among the possible diabetics we found around $1900$ samples of complications. To overcome such problem we oversampled the positives for each complication class in the training set for each experiment to achieve a better balance and avoid overfitting.

For each time gap we run a 5-fold cross-validation due to the small number of positive samples. The training of all models was implemented on PyTorch and took about 24 hours with a Nvidia Tesla P100.

\section{Results}

As baseline, we compared with a LSTM network without self-attention, but with the same pre-trained input embeddings. Table \ref{results_table} report the average area under ROC curve (AUC) over folds with the self-attentive model (LSTM+SA) and with standard LSTM.

\begin{table}[htb]
\centering
\caption{Mean AUC results of cross-validation procedure.}
\label{results_table}
\begin{center}
\begin{tabular}{c||c|c}
\multirow{2}{*}{Time Gap} &  \multicolumn{2}{c}{{\bf AUC}}\\ \cline{2-3}
&        LSTM+SA & LSTM \\ \hline
60       & 0.83  & 0.72  \\ \hline
120      & 0.82  & 0.72  \\ \hline
180      & 0.81  & 0.72  \\ \hline
240      & 0.81  & 0.71  
\end{tabular}
\end{center}
\end{table}

As the diabetic situation becomes more critical there is an increase in specialized exams and visits to the hospital\textbackslash emergency, so the performance decrease a bit as the prediction windows increase. Long term prediction seems to be a harder problem due to the lack of data.

\begin{figure}[htb]
\centering
    \includegraphics[width=\linewidth]{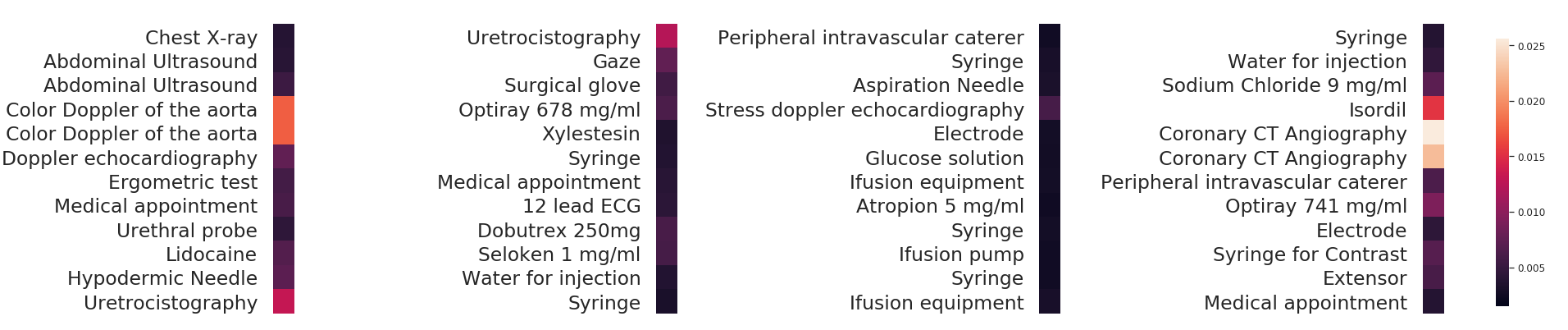}
    \centering
    \captionof{figure}{Mapping of the attention scores for an individual. Record descriptions was simplified and the latest records were organized from left to right. The individual had a angioplasty three months later.}
\label{Att_map}
\end{figure}

Comparing self-attentive model with standard LSTM we can see a better performance. This is due to the ability to deal with longer inputs with the self-attention (SA) mechanism. Figure \ref{Att_map} shows a part of neural network attention mapping scores for a particular individual. SA allowing us to see which parts of the input are attended to predict diabetes disease evolution.

\begin{figure}[htb]
\centering
\begin{minipage}{.5\textwidth}
  \centering
  \includegraphics[width=\linewidth]{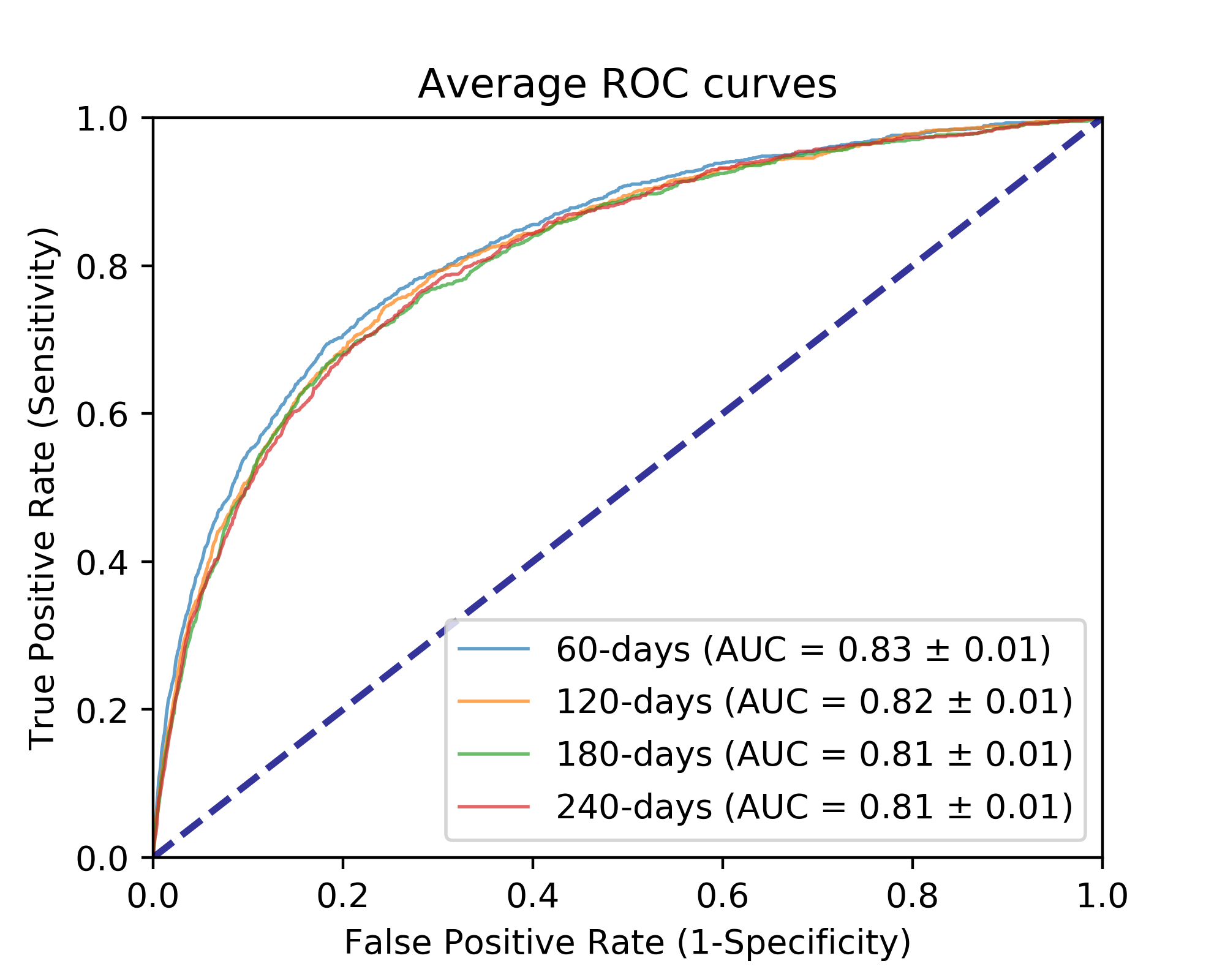}
  \captionof{figure}{Average ROC curves over time gaps}
  \label{mean_ROC}
\end{minipage}%
\begin{minipage}{.5\textwidth}
  \centering
  \includegraphics[width=\linewidth]{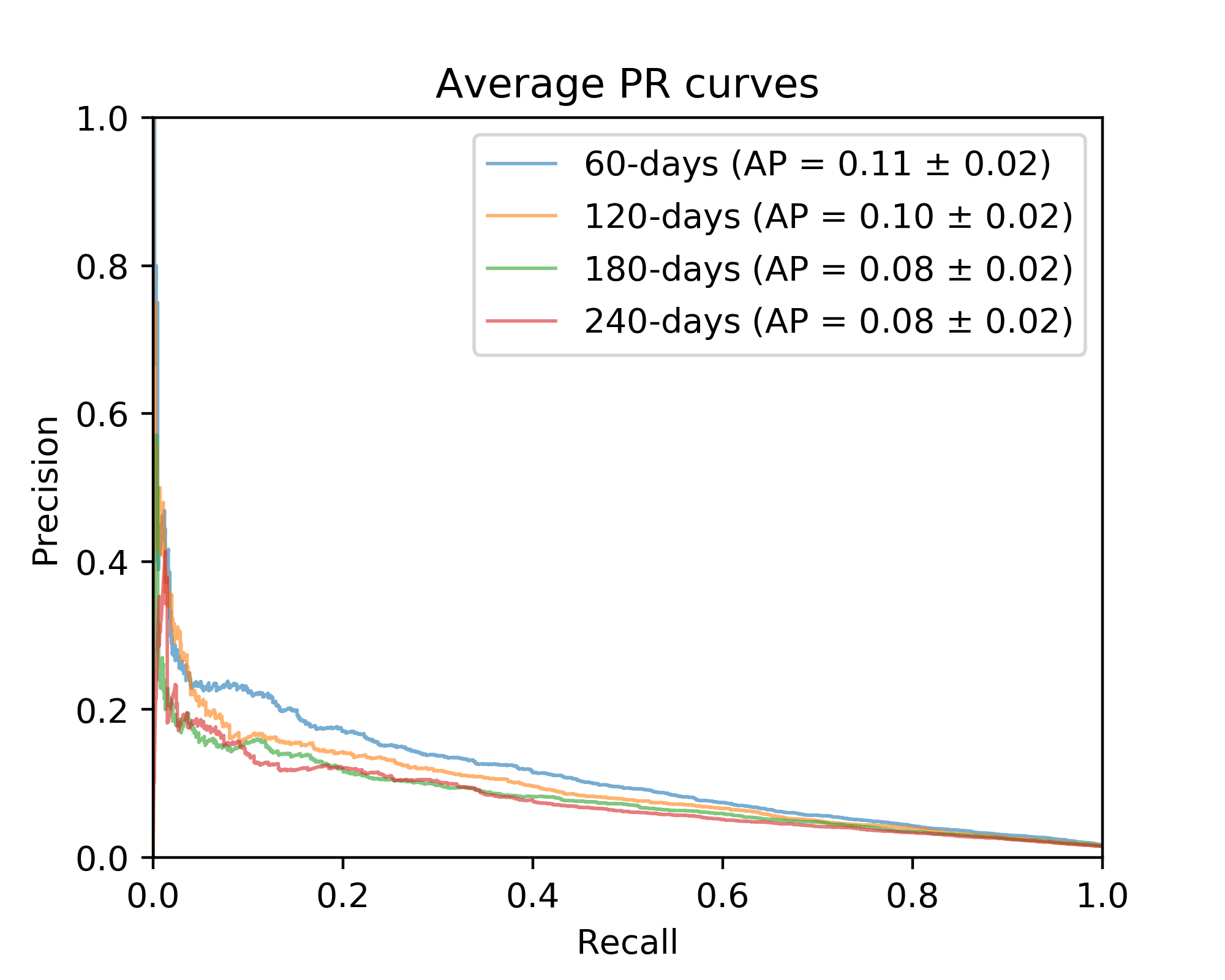}
  \captionof{figure}{Average PR curves over time gaps}
  \label{mean_PR}
\end{minipage}
\end{figure}

The average ROC and Precision-Recall curves over folds at Figures \ref{mean_ROC} and \ref{mean_PR} report the model indicates good balance between positive and negative cases and also point a limitation of how hard is to set a threshold to ensure a low false positive rate. We believe this can be related to the lack of exams results in the financial records. We inferred it is possible to detect patterns as a high frequency of relevant events, but not possible to understand the outcome of it.

Observing the data in some cases of false negative and false positive, some of those have a lack of data to do the prediction. We assumed that in some of these cases the diabetes silent progress or people refused to be treated. These indicate a challenge to increase the network specificity. One way to minimize this problem is to include these individuals in monitoring programs to give them proper treatment and then assess those with real high risk.

\section{Conclusions}
We propose the use of self-attentive recurrent neural network to predict diabetes complications using financial records. Our results demonstrated a promising methodology to predict complications and asses the high risk diabetics efficiently with an average AUC of 0.82. Despite of false positives it can still be used as an assessment tool to integrate individuals into a medium-high risk monitoring program.

At the time we are evaluating these results and studying with health plan providers the effectiveness to prevent or minimize diabetes complications. 

Using a novel method to embed dates we improved the model performance to an AUC of $0.94$. We are still validating this result and expect to publish the proposed method with a full paper report.

\bibliographystyle{unsrt}
\bibliography{nips_2018}

\clearpage

\appendix
\appendixpage
\addappheadtotoc

\begin{figure}[h]
\centering
    \includegraphics[width=1\linewidth]{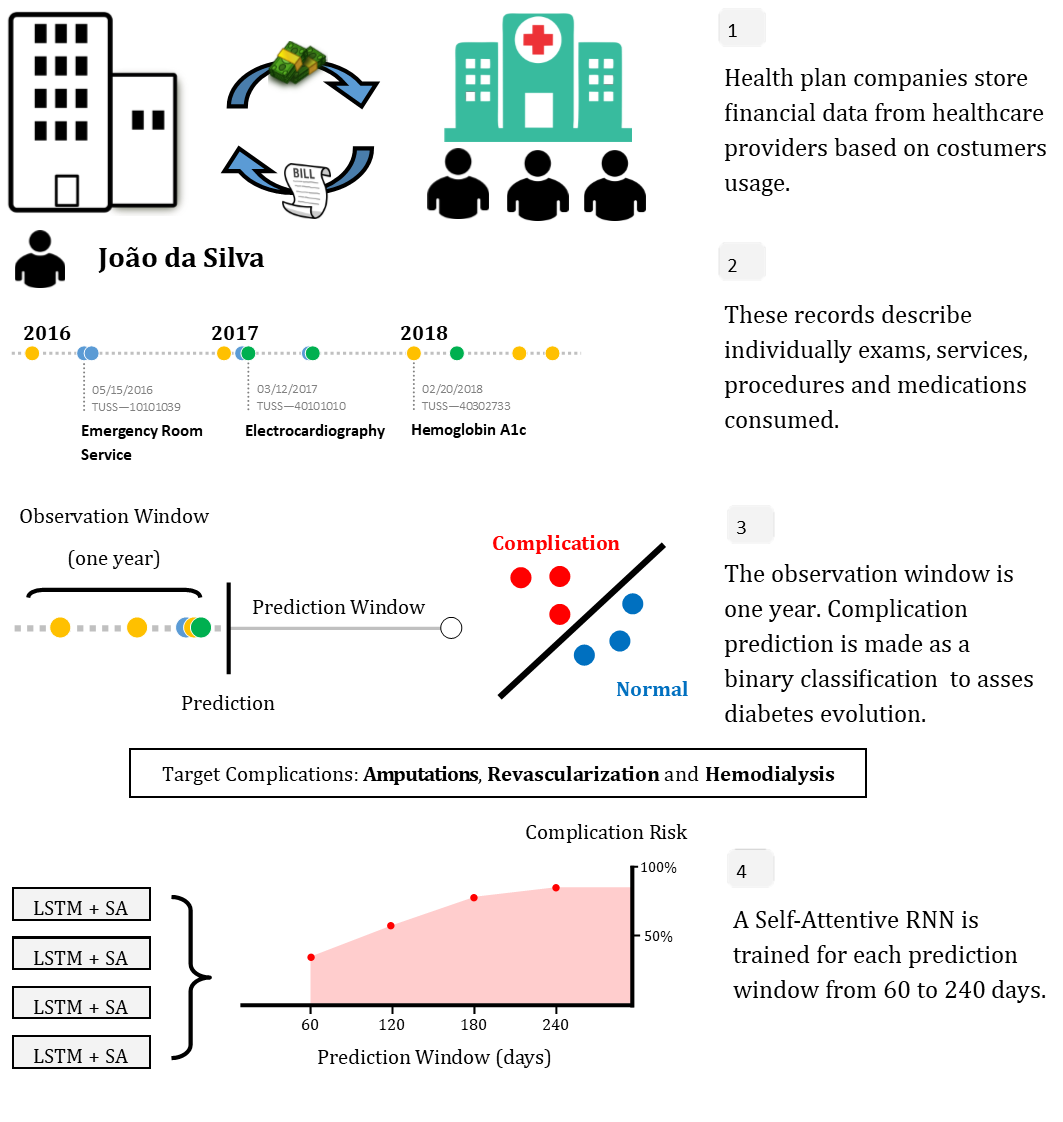}
    \centering
    \captionof{figure}{Model's workflow.}
\end{figure}

\end{document}